\documentclass{article}

\usepackage[nonatbib,final]{neurips_2024}

\usepackage[utf8]{inputenc} 
\usepackage[T1]{fontenc}    
\usepackage{hyperref}       
\usepackage{url}            
\usepackage{booktabs}       
\usepackage{amsfonts}       
\usepackage{amsmath,amssymb}
\usepackage{nicefrac}       
\usepackage{microtype}      
\usepackage{xcolor}         
\usepackage[url=false]{biblatex}
\usepackage{graphicx}
\graphicspath{ {./} }

\AtBeginBibliography{\small}

\title{GPT-2 Through the Lens of \\
Vector Symbolic Architectures}

\author{%
  Johannes Knittel\\
  Harvard University\\
   \And
   Tushaar Gangavarapu \\
   Cornell University \\
   \AND
   Hendrik Strobelt \\
   IBM Research \\
   \And
   Hanspeter Pfister \\
   Harvard University \\
}

\begin{document}

\maketitle

\begin{abstract}
  Understanding the general priniciples behind transformer models remains a complex endeavor.
  Experiments with probing and disentangling features using sparse autoencoders (SAE) suggest that these models might manage linear features embedded as directions in the residual stream. 
  This paper explores the resemblance between decoder-only transformer architecture and vector symbolic architectures (VSA) and presents experiments indicating that GPT-2 uses mechanisms involving nearly orthogonal vector bundling and binding operations similar to VSA for computation and communication between layers. It further shows that these principles help explain a significant portion of the actual neural weights.
\end{abstract}

\section{Introduction}

Understanding the way transformer models~\cite{vaswani_attention_2017} work remains a complex task, impacting AI interpretability, safety, and alignment, among other areas. Successes in using linear probing to steer models~\cite{arditi2024refusallanguagemodelsmediated,marks_geometry_2024} and principal component analysis experiments on hidden embeddings~\cite{marks_geometry_2024} indicate that these models might handle linear features embedded as directions in the residual stream, which is also known as linear representation hypothesis~\cite{elhage_toy_2022,park_linear_2024,jiang2024on}.
Recently, sparse autoencoders have been trained for a number of related model architectures based on the idea that hidden embeddings can be broken down into a combination of sparse feature vectors~\cite{bricken2023monosemanticity}.
One way how a linear additive model might function is by using nearly orthogonal vectors instead of arbitrary (or strictly orthogonal) directions.
In high-dimensional spaces, we can have many more nearly orthogonal vectors than strictly orthogonal ones~\cite{elhage_toy_2022}.
Vector symbolic architectures (VSA)~\cite{kleyko_vector_2022}, also known as hyperdimensional computing, leverage this principle for model construction.
For example, we can create 'bag-of-concept' vectors by summing nearly orthogonal vectors, so that individual concept vectors maintain a higher dot product with the summed vector, effectively modeling an OR conjunction.
The residual stream of transformers could represent such a collection of nearly orthogonal concept vectors to form a distributed code.
However, current evidence is limited on whether production models like GPT effectively use such feature vectors and how.
A better understanding of the inner mechanics has many implications on model interpretability and the usefulness of training sparse autoencoders, as well as on architectural design decisions.
This paper explores the resemblance between decoder-only transformer architectures and vector symbolic architectures and presents experiments indicating that GPT-2~\cite{radford2019language} uses mechanisms involving nearly orthogonal vector bundling and binding operations similar to VSA.
It further shows that these principles help explain a significant portion of the actual weights in the MLP layers.

\section{Interpreting GPT-2 Through Bundlings and Bindings}
\label{sec:gpt-vsa}

\begin{figure}
  \centering
  \includegraphics[width=\textwidth]{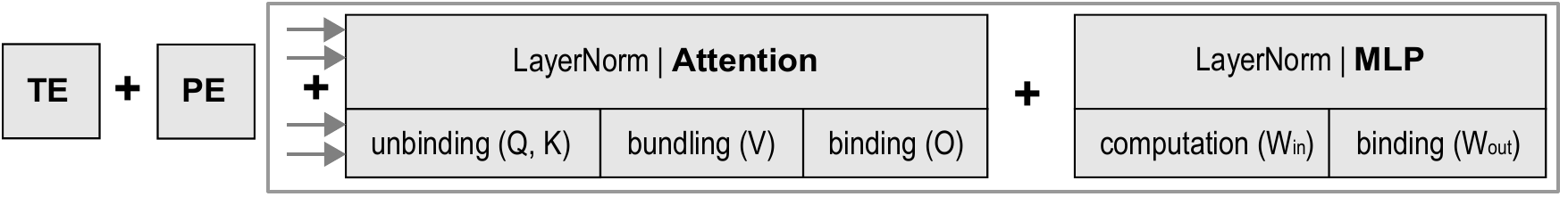}
  \caption{For each layer, the output of the attention and the feedforward network (MLP) are added to the residual stream of GPT-2, which could be interpreted as a combination of (un)binding and bundling operations using nearly orthogonal vectors.  }
  \label{fig:gpt-vsa}
\end{figure}

In high dimensional spaces, we can have many more nearly orthogonal vectors than dimensions~\cite{elhage_toy_2022}. For an $n$-dimensional space, we cannot find more than $n$ distinct vectors such that they are orthogonal to each other, i.e., the inner product (dot product for real spaces) of any vector pair is exactly zero.
However, we can find much larger sets of vectors such that the dot product of any vector to any other vector is nearly zero or small.
One can think about a unit sphere in which the surface is divided into small patches such that the centroids of these areas are sufficiently far away, corresponding to a large enough angle (i.e., small enough dot product). We can imagine that there are many more such patches than the number of dimensions, depending on our dot product threshold.
Bundling and binding of such nearly orthogonal \emph{concept vectors} are important principles of vector symbolic architectures~\cite{gallant_representing_2013}, but similar ideas are used in many other techniques, e.g., in vector-based linear models~\cite{mitchell-lapata-2008-vector}.

\emph{Bundling} refers to the bagging of vectors to create composite concept vectors.
Imagine we have three nearly orthogonal vectors $c_1, c_2, c_3$, each representing a specific (binary) concept.
We can bundle $c_1$ and $c_2$ into a concept vector $c_{1,2} = c_1 + c_2$ that has directional similarity to both vectors $c_1$ and $c_2$ but not to $c_3$.
We can efficiently test whether a vector $c_j$ is similar to either $c_1$ or $c_1$: $\langle c_{1,2}\, , c_j \rangle \geq \lambda$, essentially corresponding to a single perceptron with a ReLU activation function.

In addition to bundling, we can also perform a \emph{binding} operation by multiplying a suitable matrix $M$ with the concept vector.
For instance, this allows us to model a specific order of concepts: $cm = M  c_1 + c_2$ is similar to $c_2$, $M c_1$, and $M^{-1} c_1$, but not to $c_1$ (if $M \neq I$). $M$ could be used here to model that $c_1$ refers to the first concept of the pair $(c_1, c_2)$. 
This binding still works even if the matrix is not orthogonal but composed of nearly orthogonal vectors, i.e., $M^\intercal M$ will lead to a nearly diagonal matrix (see \autoref{app:bundling}).
 
Binding transformations can be interpreted in two ways.
We can regard them as binding operators of (possibly previously bundled) concept vectors.
We can also think of them as a construction of a new concept vector based on binary input features by adding those nearly orthogonal vectors (the rows of the matrix) where the corresponding input dimension is non-negative, with positive and negative signs as indications of the (binary) direction of the concept.
From a purely technical standpoint, both interpretations are valid.
The latter interpretation might be more helpful if we can assign semantics to specific dimensions of the input vectors (monosemanticity).

We can further build hierarchical concept vectors through bindings~\cite{wattenberg2024relational}.
For instance, if we start with a set of concepts that could apply to people in general, we might first bundle the present concepts for two persons, respectively, which we can then bind with two different matrices $M_1, M_2$. If we want to compare whether the first person in two vectors are similar, we can unbind those vectors using the inverse of $M_1$ and then take the dot product.
In other words, binding can be interpreted as a packaging step, whereas the unbinding process can be used to ‘focus’ on a specific package.
We will later outline the similarities of this interpretation to causal self attention.

Throughout this paper, we will focus on GPT-2~\cite{radford2019language} as a specific model architecture and instance of decoder-only transformers.
The flow through the hidden embeddings of transformers is often referred to as residual stream since the outputs of the attention and MLP layers are always added to the previous outputs.
This has motivated the training of sparse autoencoders (SAEs) on those hidden representations for disentangling the embeddings into a sum of sparse feature vectors, assuming that the residual stream represents many different concepts in superposition~\cite{bricken2023monosemanticity}.
Both the successful training of SAEs and experiments on finding linear representations of concepts in the embedding space are hints that transformers might store and process features using directions rather than with highly dense (and somewhat arbitrary) distributed codes.
However, sufficiently large SAEs could theoretically approximate any embedding (e.g., by dedicating one ‘feature’ for every observed training input).
Further, the sole observation of linear separability for a selection of concepts does not automatically imply that trained transformers effectively employ nearly orthogonal concept vectors and perform operations that we can interpret as matrix bindings.
On the other hand, the model architecture bears similarities with vector symbolic architectures.
We outline in this section how transformers, and GPT-2 in particular, could be interpreted with bundling and binding mechanisms closely related to VSAs.
\autoref{fig:gpt-vsa} depicts a simplified architecture of GPT-2 through the lens of VSA.

\subsection{Word Embeddings}
The word embeddings of GPT-2 are nearly orthogonal to each other, so we could interpret them as concept vectors. They are not pure or random concept vecors, though, since related tokens do point to similar directions (e.g., pairs of different versions of the ‘the’ token have high dot products). An important design decision of GPT is that the unembedding matrix at the end is just the transpose of the embedding matrix. This means that the layers need to add values to the residual stream such that the combined magnitude of those additions outweigh the initial word embedding by far. The final step in predicting the actual next token is essentially a determination which direction is the strongest since we take the dot product with every word embedding vector and then assign model probabilities through the softmax function.

\subsection{Attention Layer}
The causal self attention mechanism in GPT-2 first computes query, key, and value vectors $v_q,v_k,v_v$ by linearly transforming the current residual input $x$ with square matrices $W_Q, W_K, W_V$ (we will ignore biases for now). Let us first consider the case with just one attention head for simplicity. Then, the output of the attention added to the stream is determined by taking a weighted sum of all $v_v$ vectors of the current and previous positions, followed by a linear projection $W_O$. The weights of the sum are based on the dot product between the query and respective value vectors, normalized with the softmax function. If we assume that our learned matrices $W_Q, W_K, W_V$ are nearly orthogonal, we could interpret this process as a sequence of unbinding operations ($W_Q, W_K, W_V$), followed by the bundling of a selection of such unbound vectors $v_v$ based on their ‘relevance’, followed by the binding of the resulting sum vector through $W_O$.

In other words, the attention head focuses on a specific package of possibly bundled vectors by rotating the input space appropriately.
This package could be different for the current position (query) compared to previous positions (key).
It then computes how similar the concepts are for the specific package.
The softmax function effectively acts as a (dynamic) dot product thresholding since only those products that are sufficiently close to the maximum attention value will lead to non-zero coefficients when taking the weighted sum of the value vectors. The resulting output is bound again to obtain unique concept vectors for the specific attention layer.

Production models have more than one attention head, but this requires only minor modifications to this outlined analogy. Each attention head may be interpreted as a quantized unbinding / binding operation, that is, each head focuses on a specific package but only cares about the first few dimensions of the resulting transformation (the remaining dimensions are implicitly set to zero).
Using this interpretation, attention heads therefore copy (and repackage) concepts into the current stream, sourced from specific packages of those tokens that have conceptual similarities in specific areas.

\subsection{LayerNorm and Biases}

In GPT-2, the attention and MLP layers first apply LayerNorm before their processing (the residual stream itself is not modified, though). LayerNorm essentially centers and normalizes the input by subtracting the mean and then dividing it by the square root of the variance. It then scales the result with channel-specific learnt weights and adds a bias to each channel. We can merge the latter step (scaling and translation) into the subsequent weights and biases of the respective layer~\cite{elhage2021mathematical}. The centering and normalization depends on the actual input, though. While the normalization does not impact the angle between vectors, this is not true for translations.
Hence, accessing concepts in later layers only works if the typically observed input vectors and intermediate embeddings have a mean near zero so that distortions across layers are minimized.

The biases in $W_Q, W_K$ only add a constant factor to the dot product, which we can ignore. The biases of $W_V$ can be merged into the biases of $W_O$ as the softmax attention values always add up to one. The biases of $W_O,W_{out}$ (attention, MLP blocks) are constant additions to the residual stream. Given that they are nearly orthogonal to the word embeddings, they could serve as additional helpers for ‘cleaning’ the initial input by adding vectors that gradually decrease the share of the initial input on the final result without changing the output prediction.

\subsection{Processing of Concepts}

The MLP blocks in GPT-2 are composed of two projections. The first projects the current embedding to an intermediate vector four times the size of the hidden dimension, followed by a nonlinear activation function (ReLU or similar). The second and final projection transforms these results back to the original hidden dimension (without nonlinearities).
A single neuron in the first MLP layer could perform a simple check of whether the current stream contains a specific concept by simply performing a weighted summation in which the weights resemble the nearly orthogonal concept vector.
The nonlinearity enables more advanced boolean modeling.
An OR conjunction would be modeled by a less negative bias (presence of at least one concept contained in the weights already triggers activation), whereas an AND conjunction would require a more negative bias of said neuron (presence of two or more concepts that make up the weights are needed).
We can model the NOT operation by subtracting the respective concept vector from the weights so that they point to the opposite direction. The second layer could store the result of this operation as a new concept using a nearly orthogonal vector. With carefully selected thresholds (biases), the MLP block in transformers could therefore model boolean functions on and using nearly orthogonal concept vectors.

If individual neurons represent individual concepts, we would severely limit the number of concepts that are effectively added after each MLP block. However, as Vaintrob et al.~\cite{vaintrob_toward_2024} demonstrated for binary inputs, such processing can also result in a vector that resembles the AND conjunction of many more concepts in superposition. If we can find small sets of concepts that rarely co-occur, we can assign OR conjunctions to neurons such that they trigger whenever at least one of the concepts in the set is present. Later layers can check whether one or more concepts are actually present (AND) by testing whether all nodes containing one of said concepts have fired (i.e., all related concept vectors are present in the stream).

\section{Experiments}

The previously outlined framework is one possibility how different layers in transformers could communicate with each other and compute functions that ultimately lead to next token predictions. We performed several experiments on GPT-2 (small) to gather evidence on the suitability of this conceptual framework for the interpretation of the inner workings.
We used the TransformerLens~\cite{nanda2022transformerlens} library to extract weights and activations from GPT-2 small.
We disabled all in-built post-processing steps except for the merging of value biases into the projection biases of the attention output projection $W_{O}$ and the incorporation of the LayerNorm scaling and biases into the respective subsequent weights and biases (MLP blocks, attention blocks, and unembedding matrix).
We relied on the FAISS library~\cite{douze2024faiss} for an efficient implementation of dot product nearest neighbor search (HNSW algorithm).

\subsection{Near Orthogonality of Vectors and Matrices}

Near orthogonality of vectors and matrices are an important assumption of concept vectors and binding operations. We computed the matrix product $M^\intercal M$ for word embeddings, the attention matrices, and the output projections of the MLP blocks. \autoref{app:figures} shows some results. In all cases, we observed strong diagonals with minimal artifacts, suggesting that these matrices are indeed composed of nearly orthogonal vectors.
The output biases of the attention and MLP blocks are nearly orthogonal to the word embeddings. The magnitude of the attention output bias in the last layer is significantly higher, suggesting that this is a final cleaning step to ensure that the initial token embedding does not play a role anymore in the unembedding step.
The mean of the word embeddings and intermediate hidden embeddings for a number of prompts are near zero ($< 0.05$), suggesting low distortions of the angles of directions across layers.

\subsection{Processing of Concepts}

\begin{table}
  \caption{Explaining neurons by comparing their weights with sums of token embeddings.}
  \label{tab:neuralExplanations}
  \centering
  \begin{tabular}{lll}
    \toprule
    Neuron     & Vector     & Cos. Sim. \\
    \midrule
    0-12 & \small{ 

\texttt{\colorbox{yellow!30}{  CNN}} $+$ \texttt{\colorbox{yellow!30}{  Cold}} $+$ \texttt{\colorbox{yellow!30}{  NBC}} $+$ … $+$ \texttt{\colorbox{yellow!30}{  lying}} $+$ … $+$ \texttt{\colorbox{yellow!30}{ TRUMP}} $+$  … }   & $0.61$     \\
    0-186 & \small{ \texttt{\colorbox{yellow!30}{  remarked}} $+$ \texttt{\colorbox{yellow!30}{  commented}} $+$ \texttt{\colorbox{yellow!30}{  complained}} $+$ \texttt{\colorbox{yellow!30}{  said}} $+$ … }   & $0.63$     \\
    0-192 & \small{ \texttt{\colorbox{yellow!30}{  being}} $+$ \texttt{\colorbox{yellow!30}{ Being}} $+$ \texttt{\colorbox{yellow!30}{ Having}} $+$ \texttt{\colorbox{yellow!30}{  be}} $+$ \texttt{\colorbox{yellow!30}{ Be}} $+$ \texttt{\colorbox{yellow!30}{  beings}} $+$  … }   & $0.65$     \\
    0-205 & \small{ \texttt{\colorbox{yellow!30}{  selections}} $+$ \texttt{\colorbox{yellow!30}{  eclectic}} $+$ \texttt{\colorbox{yellow!30}{  selection}} $+$ \texttt{\colorbox{yellow!30}{  assortment}}  $+$ … }   & $0.58$     \\
    0-247 & \small{ \texttt{\colorbox{yellow!30}{  resulted}} $+$ \texttt{\colorbox{yellow!30}{  based}} $+$ \texttt{\colorbox{yellow!30}{ Given}} $+$ \texttt{\colorbox{yellow!30}{  highlighted}}  $+$ … }   & $0.62$     \\
    \bottomrule
  \end{tabular}
\end{table}

Earlier experiments indicated that some neurons in GPT-2 focus on very specific tokens, such as \textit{the}. Assuming that MLP blocks perform boolean operations on concept vectors and that word embeddings are concept vectors, we would expect some neurons to have weights representing either a single word vector or a composite vector made of several word vectors (e.g., the sum of all word embeddings that are variations of \textit{the}).
To test this hypothesis, we conducted experiments.

Specifically, we aimed to determine if some neural weights could be explained by \emph{simple} bundled vectors that are merely sums of word embeddings.
For each possible input vector (word embedding), we compiled a list of the most similar neurons.
That is, we looked for neurons in the first feedforward layer of the MLP block whose input weights have a reasonably high dot product with the centered candidate vector.
For each neuron and its list of candidates, we then greedily assembled the bundled vector by including those vectors (ranked by similarity) that lead to an increase of the cosine similarity between the bundled vector and the respective neural weights.
We discarded vectors with a cosine similarity of less than $0.05$ with the neural weight vector.
We also discarded those with less than $0.1$ if they did not significantly increase the overall cosine similarity by more than $0.04$.
These thresholds represent hyperparameters and were conservatively chosen to obtain sets of highly relevant vectors.
We enforced a minimal similarity to the target vector and only considered unweighted sums since any vector can be trivially represented by a weighted sum of sufficiently many nearly orthogonal vectors.

With this simple approach, we achieve a similarity of $0.5$ or higher for more than 80\% of the neurons in the first layer, and at least $0.3$ for more than 95\% (this includes attributions to attention heads we discuss in the next section).
For instance, our greedily obtained vector for neuron 1844 in the first layer has a high cosine similarity of $0.69$ with the actual weights of that neuron. The list contains many first names, most of them male sounding (' Chris', ' Kevin', ' Jeff') with some exceptions (' Rebecca'). Neuronpedia explains this as \textit{first names of people}. 
Neuron 20 (similarity of $0.73$) focuses on tokens somehow semantically related to 'maintaining', including 'Nevertheless' and 'Storage' (Neuronpedia says \textit{verbs related to maintaining or keeping something}).
\autoref{tab:neuralExplanations} lists more examples.
It is important to note that we inferred these neuron explanations directly from the weights rather than through activation observations and correlations, or by training an autoencoder or probing classifier.

\subsection{Processing Circuits}
\begin{figure}
  \centering
  \includegraphics[width=0.9\textwidth]{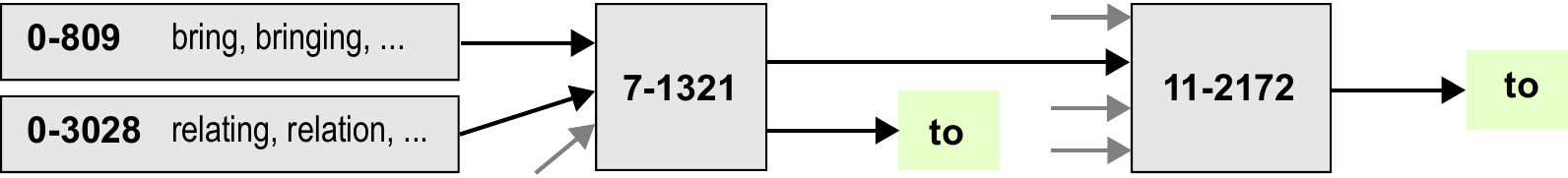}
  \caption{Circuit discovery using similarities between $W_{out}$ and $W_{in}$ vectors based on greedily obtained concept vectors. The two neurons in the first layer each focus on a set of tokens. The neuron 7-1321 seems to model a union of those sets, whose output is similar to the input of the depicted neuron in the last layer and the \textit{to} unembedding. The last one has many inbound connections to previous outputs and seems to boost the concept of the \textit{to} vector.  }
  \label{fig:circuit}
\end{figure}

Using the analogy of concept vectors and matrix bindings, we would expect that the MLP blocks perform (boolean) functions on concept vectors sourced from either the token embeddings, previous results from MLP blocks (information from computations), or attention blocks (information from other tokens). The results are then written back to the stream via matrix binding.
The sum of all written concept vectors (including those from the attention blocks) should exhibit the highest directional similarity with the token embedding that is up next. This led us to investigate whether neural weights could be explained based on \( W_O \) vectors (output projection vectors of the attention layer) and \( W_{out} \) vectors (final projection in MLP block).
This would support the hypothesis that GPT-2 communicates via near-orthogonal vectors between layers. This communication either aims to bring the stream closer to the predicted next token or serves as an input for subsequent circuits.
Analogous to the previous experiment, we tried to explain neural weights with simple bundled vectors.
For this set of experiments, we included output projection vectors of the attention layer \( W_O \) and of the MLP block \( W_{out} \), in addition to the word embeddings.

We were able to reconstruct bundled vectors with a cosine similarity of at least $0.3$ with neural input weights in 30-50\% of neurons within the middle layers, as well as the majority of neurons in both the first and last layer.
The majority of bundlings were composed of up to 40 vectors.
For some neurons, our simple circuit discovery strategy explained the weights nearly perfectly.
For instance, neuron 2537 in layer 1 can be explained by the non-activation of neurons 2977 and 1993 in layer 0, showing a similarity of $0.91$.
Neuron 1073 in layer 1 seems to be dependent on the presence of concept 11 in head 1.4 and the absence of concepts 31 in head 1.2 and 1 in head 1.10, with a weight similarity of $0.72$.
\autoref{fig:circuit} illustrates a circuit that seems to increase the likelihood of predicting the word \textit{to} following certain words like \textit{bring} or \textit{relating}.
When forcing the output of neuron 7-1321 to zero, the logit for the token \textit{to} decreases for the prompt \textit{In relation} but not for the prompt \textit{I was listening}.
Neuron 12 in the first layer appears to trigger on a specific set of tokens including names of major news outlets (e.g., CNN) and the word \textit{TRUMP}. Using our strategy, we were unable to identify other neurons significantly influenced by the output of this neuron. However, we observed a cosine similarity of 0.07 with the token \textit{Greenwald}. Although this seems low, it is crucial to remember that tokens potentially encompass a broad array of concepts. The model may have overfitted to news articles related to Glenn Greenwald and the US election.

The first and last layers have strong similarities to token embeddings.
The residual stream in higher layers already points toward likely tokens that should be predicted next. One possible explanation is that the last layer therefore weakens or confirms current predictions by also taking into account which tokens are likely to be predicted next.
It is important to note that we employed a greedy method for identifying related vectors and thus circuits, which assumes a certain degree of monosemanticity in individual neurons and attention dimensions.
Our objective was to gather evidence of matrix binding and the processing of nearly orthogonal vectors, rather than to identify all communication channels comprehensively.

\section{Related Work}

Linear representations in models and embeddings have a long history in research, with Word2Vec~\cite{mikolov2013distributed} popularizing the idea of performing simple arithmetic on word embeddings.
Numerous probing experiments~\cite{alain2017understanding} have been performed to investigate linear representations in large language models~\cite{tenney2018what,tenney2019bert,gurnee_finding_2023,li2023emergent,arditi2024refusallanguagemodelsmediated,marks_geometry_2024}.
Hernandez et al.~\cite{hernandez2024linearity} found evidence that relational decoding in large language models can often be approximated with an affine transformation.
Wattenberg and Vegas~\cite{wattenberg2024relational} discuss relational composition in context of sparse autoencoders, noting that matrix bindings bear similarities with the attention process.
The field of mechanistic interpretability~\cite{olah2020zoom} is concerned with reverse engineering and explaining (in parts) the inner workings of models, for instance, through discovering and describing circuits~\cite{hanna2023how,wang2023interpretability}, or by trying to understand whether certain behavior can be tracked down to individual neurons or is represented in a more distributed way~\cite{dalvi_what_2018,bau_understanding_2020,durrani_analyzing_2020,elhage_toy_2022,gurnee_finding_2023}.

\section{Discussion and Conclusion}

Our experiments indicate that GPT-2 effectively learns mechanisms that are similar to concepts of vector symbolic architectures.
The different blocks seem to communicate by writing and reading nearly orthogonal vectors from the residual stream.
We could also show that at least some neurons implement simple boolean functions as outlined in Section~\ref{sec:gpt-vsa}.
This has several implications on model interpretability and architectural design decisions.
The presence of bundling and binding circuits strengthens the validity of training sparse autoencoders for model disentanglement.
However, such nearly orthogonal vectors may not necessarily map to interpretable features or independently understandable features (decomposability theory~\cite{elhage_toy_2022}).
They may be part of what Engels et al. describe as multidimensional features~\cite{engels2024not}.
Our findings may also inform future architectural decisions. For instance, it may help to understand which architectural components promote or hinder the learning of bundling and binding operations, possibly leading to better model architectures.
A better understanding of the inner workings also helps to hypothesize for which type of tasks GPT-like models work better or worse.

We want to emphasize that our findings are preliminary and our method has several limitations.
We only performed experiments on GPT-2 small and further experiments are needed to confirm that these behaviors also exist in related and more recent models. We deliberately focused on very simple techniques and avoided the use of sparse autoencoders in our experiments to mitigate confounding factors and uncertainties introduced by additional modeling.

\begin{ack}
This work was partially supported by the Harvard Data Science Initiative Postdoctoral Fellowship.
\end{ack}

\printbibliography


\appendix

\section{Bundling and Binding of Concept Vectors}
\label{app:bundling}

In this paper, we will consider real valued space $\mathbb{R}^n$ with $n \gg 100$. In such high dimensional spaces, we can have many more nearly orthogonal vectors than dimensions~\cite{elhage_toy_2022}. For an $n$-dimensional space, we cannot find more than $n$ distinct vectors such that they are orthogonal to each other, i.e., the inner product (dot product) of any vector pair is exactly zero.
However, we can find much larger sets of vectors such that the dot product of any vector to any other vector is nearly zero or small.
One can think about a unit sphere in which the surface is divided into small patches such that the centroids of these areas are sufficiently far away, corresponding to a large enough angle (i.e., small enough dot product). We can imagine that there are many more such patches than the number of dimensions, depending on our dot product threshold.

Imagine we have three big vectors $c_1, c_2, c_3$, each representing a specific (binary) concept.
If these vectors are nearly orthogonal, we can create a compositional concept vector $c_{1,2} = c_1 + c_2$ that has directional similarity to both vectors $c_1$ and $c_2$ but not to $c_3$.
In vector symbolic architectures (VSA), this is referred to as bundling or superpositioning.
We can test whether a vector $c_j$ is similar to either $c_1$ or $c_1$ with a simple dot product and thresholding, i.e., $\langle c_{1,2}\, , c_j \rangle \geq \lambda$. 
This summation essentially represents an OR conjunction (‘bag of concepts’ vector). 
Please note the striking similarity of this test to a single perceptron with a ReLU activation function.

In addition to summation, we can also perform a binding operation by multiplying a suitable invertible matrix $M$ with the concept vector.
For instance, this allows us to model a specific order of concepts: $cm = M  c_1 + c_2$ is similar to $c_2$, $M c_1$, and $M^{-1} c_1$, but not to $c_1$ (if $M \neq I$). $M$ could be used here to model that $c_1$ refers to the first concept of the pair $(c_1, c_2)$. We can see that if $M$ is chosen well, the dot products are preserved:
$\langle M c_1\, , M c_2 \rangle =  ( M c_1 )^\intercal ( M c_2 ) = c_1^\intercal M^\intercal M c_2$

If $M$ is an orthogonal matrix, this corresponds to a basis transformation, preserving the length and dot products of vectors as it is just a rotation (or reflection) in simpler terms and $M^\intercal M$ is just the identity matrix. 
Even if the matrix is not orthogonal but composed of nearly orthogonal vectors, $M^\intercal M$ will lead to a nearly diagonal matrix. Looking at the inner product definition, we can see that if the scalars are roughly similar, the angle between vectors does not change much after the binding transformation.
We will call such matrices nearly orthogonal matrices given that they preserve the dot product sufficiently well. 
 
Such transformations can be interpreted in two ways.
We can regard them as binding operators of (possibly previously bundled) concept vectors.
We can also think of them as a construction of a new concept vector based on binary input features by adding those nearly orthogonal vectors (the rows of the matrix) where the corresponding input dimension is non-negative, with positive and negative signs as indications of the (binary) direction of the concept.
From a purely technical standpoint, both interpretations are valid.
The latter interpretation might be more helpful if we can assign semantics to specific dimensions of the input vectors (monosemanticity).

We can build hierarchical concept vectors through bindings~\cite{wattenberg2024relational}.
For instance, if we start with a set of concepts that could apply to people in general, we might first take two sums capturing the present concepts of two specific persons, respectively, which we can then bind with two different matrices $M_1, M_2$. If we want to compare whether the first person in two vectors are similar, we can unbind those vectors using the inverse of $M_1$ and then take the dot product.
In other words, matrix binding can be interpreted as a packaging step, whereas the unbinding process can be used to ‘focus’ on a specific package.
Bundling and binding of nearly orthogonal vectors are important principles of vector symbolic architectures~\cite{gallant_representing_2013}, but similar ideas are used in many other techniques, e.g., in vector-based linear models~\cite{mitchell-lapata-2008-vector}.

\section{Figures}
\label{app:figures}

\begin{figure}
  \centering
  \includegraphics[width=0.9\textwidth]{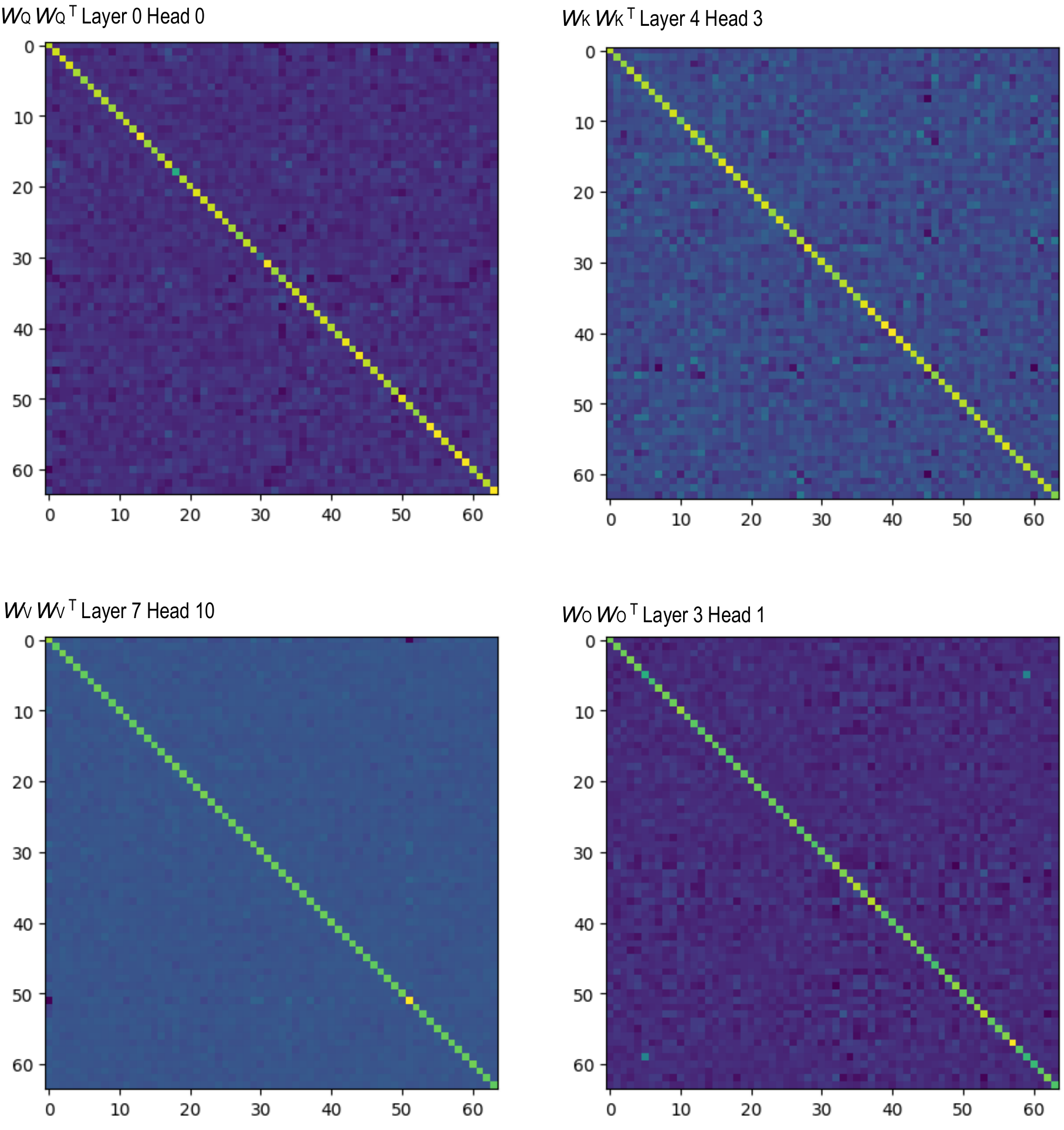}
  \caption{The attention matrices are composed of nearly orthogonal vectors. This is an exemplary selection of attention heads and their corresponding matrix multiplication $M M^\intercal$. }
  \label{fig:attention-matrices}
\end{figure}

\begin{figure}
  \centering
  \includegraphics[width=0.9\textwidth]{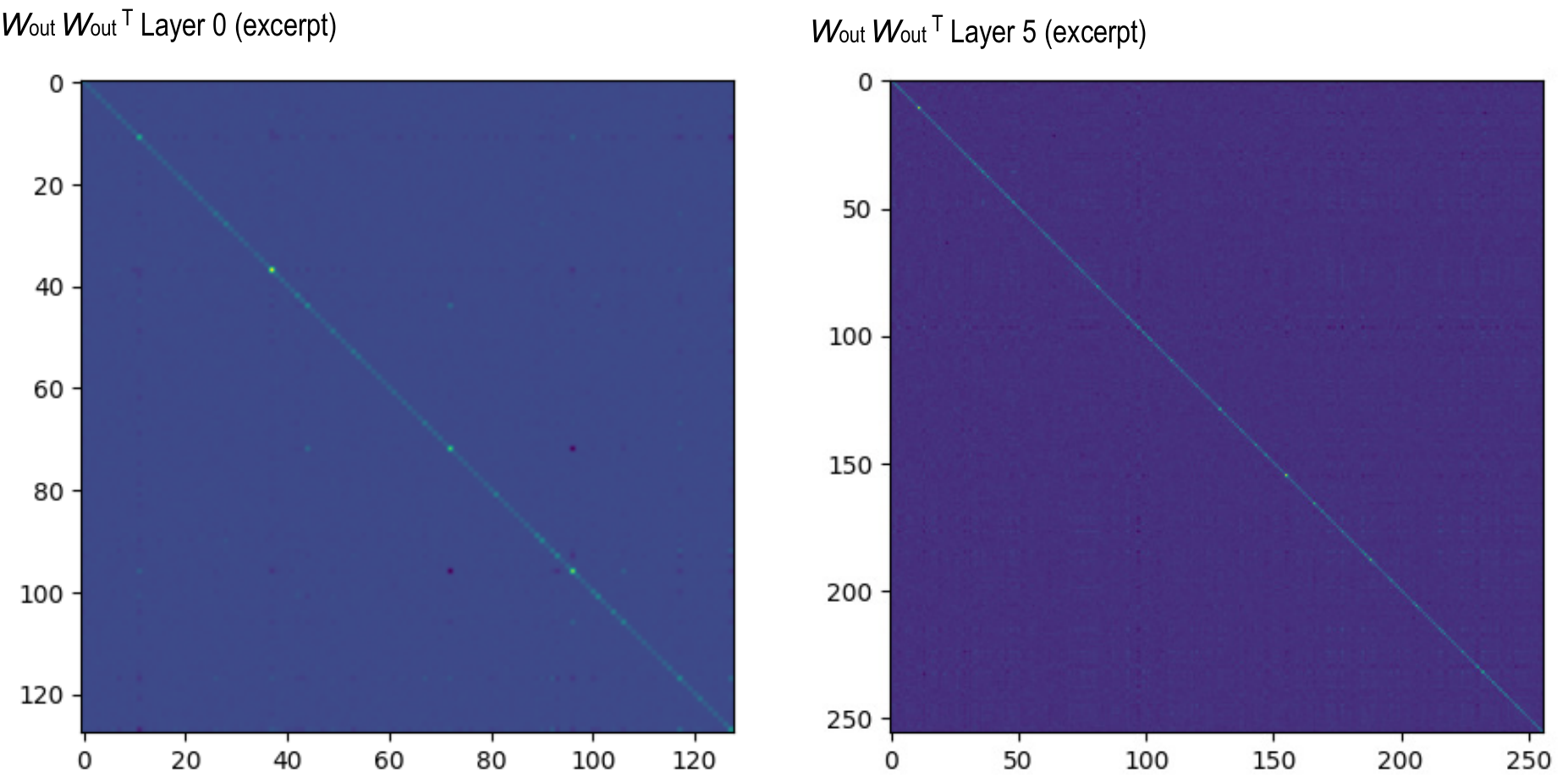}
  \caption{The output projection matrices of the MLP blocks are composed of nearly orthogonal vectors. This is an exemplary selection of output projection matrices for specific layers and the corresponding matrix multiplication $M M^\intercal$. We only show the top left cutout of the visualization as each layer contains 3072 neurons. }
  \label{fig:wout-matrices}
\end{figure}

Near orthogonality of vectors and matrices are an important assumption of concept vectors and binding operations. We computed the matrix product $M^\intercal M$ for word embeddings, the attention matrices, and the output projections of the MLP blocks. In all cases, we observed strong diagonals with minimal artifacts, suggesting that these matrices are indeed composed of nearly orthogonal vectors.
\autoref{fig:attention-matrices} and \autoref{fig:wout-matrices} show exemplary visualizations of a selection of attention heads and MLP layers.

\end{document}